\documentclass{article} 
\usepackage{iclr2026_conference,times}

\usepackage{amsmath,amsfonts,bm}

\def\eqref#1{equation~\ref{#1}}

\def\1{\bm{1}}

\DeclareMathAlphabet{\mathsfit}{\encodingdefault}{\sfdefault}{m}{sl}
\SetMathAlphabet{\mathsfit}{bold}{\encodingdefault}{\sfdefault}{bx}{n}

\usepackage{hyperref}
\usepackage{url}
\usepackage[T1]{fontenc}
\usepackage[utf8]{inputenc}

\usepackage{graphicx}
\usepackage{pgfplots}
\pgfplotsset{compat=1.18} 
\title{ATLAS: Benchmarking and Adapting LLMs for Global Trade via Harmonized Tariff Code Classification}

\author{
Pritish Yuvraj \\
Flexify.AI \\
\texttt{pritish@flexify.ai} \\
\And
Siva Devarakonda\thanks{Website: \url{https://tariffpro.flexify.ai/}} \\
Flexify.AI \\
\texttt{siva@flexify.ai} \\
}
\iclrfinalcopy 
\begin{document}

\maketitle
\begin{abstract}
Accurate classification of products under the Harmonized Tariff Schedule (HTS) is a critical bottleneck in global trade, yet it has received little attention from the machine learning community. Misclassification can halt shipments entirely, with major postal operators suspending deliveries to the U.S. due to incomplete customs documentation.  

We introduce the first benchmark for HTS code classification, derived from the U.S. Customs Rulings Online Search System (CROSS). Evaluating leading LLMs, we find that our fine-tuned \textsc{Atlas} model (LLaMA-3.3-70B) achieves \textbf{40\% fully correct 10-digit classifications} and \textbf{57.5\% correct 6-digit classifications}—improvements of \textbf{+15 points over GPT-5-Thinking} and \textbf{+27.5 points over Gemini-2.5-Pro-Thinking}.  

Beyond accuracy, \textsc{Atlas} is \textbf{5$\times$ cheaper than GPT-5-thinking} and \textbf{8$\times$ cheaper than Gemini-2.5-Pro-Thinking}, and can be self-hosted to guarantee data privacy—an essential requirement in high stake industries like Automotives ,Industrials, Semiconductors etc. for trade and compliance workflows. While \textsc{Atlas} sets a strong baseline, the benchmark remains highly challenging, with only 40\% 10-digit accuracy.  

By releasing both dataset and model, we aim to position HTS classification as a new community benchmark task. We invite future work in retrieval, reasoning and alignment to advance progress on this high-impact global trade problem.
\end{abstract}

\footnote{\scriptsize 
1. HTS CROSS Rulings Dataset: \url{https://huggingface.co/datasets/flexifyai/cross_rulings_hts_dataset_for_tariffs} \\
2. Atlas LLM Model: \url{https://huggingface.co/flexifyai/atlas-llama3.3-70b-hts-classification}

3. Atlas LLM Model Demo: \url{https://flexifyai-atlas-llama3-3-70b-hts-demo.hf.space/?logs=container&__theme=system&deep_link=FFJuTJsv_fM}
}

\section{Introduction}

Every product imported into the global market must be assigned a Harmonized Tariff Schedule (HTS) code. 
These codes, standardized by the World Customs Organization (WCO), are ten digits long. 
The first six digits are harmonized across all participating countries, while the last four digits are country-specific extensions. 
Correctly identifying the first six digits enables global interoperability, while the full ten-digit code is required for compliance with U.S. customs.  

The HTS is deeply hierarchical: 22 sections are divided into 99 chapters, which expand into thousands of subheadings. 
Chapters 1–97 correspond to stable product categories (such as chemicals, machinery, and textiles), whereas Chapters 98–99 capture temporary and special provisions that change frequently. 
This structure makes tariff classification a natural hierarchical machine learning task, where six-digit accuracy captures worldwide consistency, and ten-digit accuracy reflects the U.S.-specific extension.

Despite its centrality, classification remains a major bottleneck. 
Recent trade policy changes, for example, the modifications to the \emph{de minimis} exemption, require that any imported good valued above \$100 must be assigned a valid HTS code. 
The HTS itself spans more than 17,000 pages of PDF documents, making manual assignment infeasible at scale. 
The consequences are global: in 2025, several major postal operators suspended parcel delivery to the United States, citing the inability to assign correct HTS codes and complete customs documentation \cite{timesofindia2025htsban}, \cite{reuters2025germanpost}, \cite{usatoday2025postalban}. 
More than thirty countries were affected, highlighting the fragility of global trade flows when classification is not available at scale.

Large Language Models (LLMs) offer a scalable alternative. 
Their capacity for semantic reasoning and structured classification makes them natural candidates for HTS code classification, where fine-grained distinctions must be captured. 
Moreover, since the first six digits are harmonized worldwide, advances in HTS classification can simultaneously benefit global trade systems, while the U.S.-specific digits directly address compliance in the American market.

\subsection{Contributions}
Our key contributions are:  

\begin{itemize}
    \item We release the first open-source benchmark for HTS classification \cite{flexify2025huggingface}, constructed from the U.S. Customs Rulings Online Search System (CROSS), including training, validation, and test splits.  
    \item We benchmark leading proprietary and open-source models, including GPT-5-Thinking \cite{openai2025gpt5}, Gemini-2.5-Pro-Thinking \cite{deepmind2025gemini25pro}, LLaMA-3.3-70B \cite{grattafiori2024llama3herdmodels}, DeepSeek-R1 (05/28) \cite{deepseekai2025deepseekr1incentivizingreasoningcapability}, and GPT-OSS-120B \cite{openai2025gptoss120bgptoss20bmodel}.  
    \item We fine-tuned LLaMA-3.3-70B with supervised fine-tuning (SFT) to create the specialized model \textsc{Atlas}, which we open source~\cite{yuvraj2025atlas}. \textsc{Atlas} achieves \textbf{40\% accuracy at the 10-digit level} and \textbf{57.5\% at the 6-digit level}, substantially outperforming GPT-5-Thinking (25\%) and Gemini-2.5-Pro-Thinking (13.5\%).  
    \item Beyond accuracy, \textsc{Atlas} is significantly more cost-efficient—up to \textbf{8$\times$ cheaper per inference}—and supports privacy-preserving deployment through self-hosting, ensuring that sensitive trade data never leaves secure environments.  
\end{itemize}

Together, these contributions establish tariff code classification as a new frontier for LLM evaluation, situated at the heart of compliance for Global Commerce and Trade.

\subsection{Paper Roadmap}
The remainder of this paper is organized as follows. 
Section~\ref{gen_inst} describes the construction of our CROSS-based dataset and its transformation into a machine-learning–ready format. 
Section~\ref{sec:model_training} details the fine-tuning procedure for \textsc{Atlas}. 
Section~\ref{sec:results} presents evaluation results across multiple proprietary and open-source LLMs, analyzing both accuracy and cost efficiency. 
Finally, we conclude with a summary of key findings and future research directions in Section~\ref{summary}.

\section{Datasets}
\label{gen_inst}

A central contribution of this work is the construction of the first large-scale dataset for Harmonized Tariff Schedule (HTS) classification, derived from the U.S. Customs Rulings Online Search System (CROSS) \cite{cbp2025rulingsonline}. CROSS contains legally binding decisions issued by U.S. Customs and Border Protection (CBP), in which importers or brokers sought clarification on the correct HTS code for specific products. These rulings are authoritative, high-value examples of tariff classification in practice, but are lengthy, unstructured, and scattered across thousands of HTML pages—making them inaccessible for machine learning research.  

\subsection{Data Collection}
We developed an automated browser agent \cite{selenium,chromedriver,webdriver_manager} to systematically scrape CROSS. Each ruling was matched to a 10-digit HTS code obtained from the official HTS U.S. website \cite{usitc2025hts}. After filtering and cleaning, the final dataset spans 18{,}731 rulings covering 2,992 unique HTS codes across a broad range of product categories.  

Not every HTS code appears in CROSS, since only disputed or clarified codes are documented. However, the presence of a code in CROSS is itself informative: frequent rulings signal categories that are high-demand or ambiguous in practice, while absent codes suggest stable or rarely used classifications.   

\subsection{Data Transformation into LLM-Trainable Format}
Raw CROSS rulings are official letters—legalistic, verbose, and inconsistent in structure. To make them suitable for supervised learning, we transformed each ruling into a structured prompt–response format using GPT-4o-mini \cite{openai2024gpt4ocard}. This lightweight model was cost-effective and sufficient for information extraction.  

\paragraph{Prompt template.} Each ruling was converted into the following instruction format:

\begin{verbatim}
Given the following HTS ruling information:

HTS Code: {hts_code}
Ruling Number: {ruling_number}
Title: {title}
Date: {date}
URL: {url}
Summary: {summary}
Content: {content}

Please analyze this information and provide:

a) Create a product description that the user was initially getting the HTS US code for
b) Create a reasoning path justifying why the HTS US code is correct
c) Return the HTS US code

Format your response as follows:

User: What is the HTS US Code for [product_description]?
Model:
HTS US Code -> [HTS US Code]
Reasoning -> [detailed_reasoning_path]
\end{verbatim}

This design forces models to both predict the code and provide a reasoning path, aligning with recent work on chain-of-thought reasoning \cite{wei2023chainofthoughtpromptingelicitsreasoning}.  

\subsection{Dataset Splits}
From the 18{,}731 processed rulings, we randomly sampled 200 examples for validation and 200 for final testing, holding them out strictly from training. The remaining 18{,}254 rulings form the training set. This ensures a clean separation between model development and final evaluation. We have uploaded the dataset to hugging-face \cite{flexify2025huggingface}.

\begin{table}[h]
\centering
\caption{Distribution of the CROSS dataset into training, validation, and test splits.}
\label{tab:dataset_splits}
\begin{tabular}{|l|c|}
\hline
\textbf{Split} & \textbf{Number of Rulings} \\
\hline
Training   & 18{,}254 \\
Validation & 200 \\
Test       & 200 \\
\hline
\end{tabular}
\end{table}

\subsection{Discussion}
This dataset poses unique challenges: (1) rulings are lengthy and often hinge on subtle distinctions (e.g., partially fabricated vs. fully fabricated semiconductor wafers); (2) correctness has a hierarchical structure (6-digit vs. 10-digit); and (3) errors carry real-world consequences for trade and compliance. By releasing this benchmark, we aim to establish tariff classification as a novel, high-impact evaluation task for LLMs, complementing existing benchmarks in reasoning, code generation, and multilingual understanding.
\section{Model Training}
\label{sec:model_training}

While several open-source large language models could, in principle, be adapted for tariff classification, we made a deliberate and principled choice to focus exclusively on \textbf{LLaMA-3.3-70B} \cite{grattafiori2024llama3herdmodels}. Two factors motivated this decision. First, practical \emph{budget constraints} made it infeasible to fine-tune multiple frontier models at scale. Second, LLaMA-3.3-70B is a dense architecture, making it both simpler to fine-tune and easier to deploy in inference settings compared to Mixture-of-Experts (MoE) architectures such as DeepSeek-R1 or GPT-OSS-120B. From a community perspective, providing a dense and reproducible baseline lowers the entry barrier for downstream research: training and inference pipelines are easier to set up, memory usage is more predictable, and accuracy is less sensitive to expert routing heuristics.  

\subsection{Supervised Fine-Tuning Objective}
We adapted LLaMA-3.3-70B to the CROSS dataset using supervised fine-tuning (SFT) \cite{brown2020language,ouyang2022training}. Each ruling was transformed into an input–output pair, where the input is a ruling-derived product description and the output is the correct HTS code along with a reasoning trace. This makes the task well aligned with the SFT paradigm, which minimizes the token-level negative log-likelihood of ground-truth outputs.  

Formally, for an input sequence $x=(x_1,\dots,x_n)$ and target sequence $y=(y_1,\dots,y_m)$, the model with parameters $\theta$ defines conditional probabilities $p_\theta(y_t \mid x, y_{<t})$. The training loss is then:  

\[
\mathcal{L}_{\text{SFT}}(\theta) = - \sum_{t=1}^{m} \log p_\theta(y_t \mid x, y_{<t}),
\]

which corresponds to the standard negative log-likelihood objective.  

\subsection{Training Setup and Stability}
Fine-tuning was performed for 5 epochs (approximately 1,400 steps) using the AdamW optimizer with $\beta_1=0.9$, $\beta_2=0.95$, weight decay $=0.1$, and a cosine learning-rate schedule initialized at $1 \times 10^{-7}$. To manage the high memory footprint of 70B-parameter models, we employed bf16 precision and gradient accumulation to simulate a batch size of 64 sequences. Training was distributed across 16 $\times$ A100-80GB GPUs using fully sharded data parallelism.  

\begin{figure}[h]
    \centering
    \includegraphics[width=0.6\linewidth]{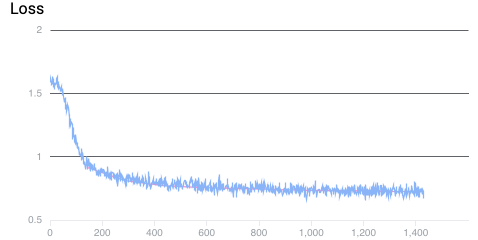}
    \caption{Training loss curve over 1,400 optimization steps. Rapid early improvement is followed by stable convergence.}
    \label{fig:loss_curve}
\end{figure}

As shown in Figure~\ref{fig:loss_curve}, the training loss decreases sharply in the first 200 steps and then stabilizes near convergence, with no sign of overfitting on the validation set. We observed stable gradient norms and no catastrophic spikes in loss, suggesting that dense models like LLaMA-3.3-70B are well suited to small but domain-specific datasets when carefully regularized. This highlights that reproducible fine-tuning of frontier models is feasible even under modest compute budgets, provided that optimization choices are tuned to stability.  

\subsection{Ablations and Future Work}
While our study focused exclusively on LLaMA-3.3-70B, several ablation studies could provide deeper insights and further guide the community:  

\begin{itemize}
    \item \textbf{Model scale:} Evaluating smaller LLaMA variants (e.g., 8B or 3B) would clarify the tradeoff between accuracy, cost, and deployability on edge devices.  
    \item \textbf{Retrieval augmentation:} Integrating retrieval over the 17,000-page HTS documents may reduce hallucinations and improve long-tail classification accuracy, complementing SFT.  
    \item \textbf{Contrastive and hybrid objectives:} Beyond NLL, contrastive learning between closely related codes (e.g., semiconductor wafers vs. finished chips) may sharpen decision boundaries.  
    \item \textbf{Direct Preference optimization:} Beyond NLL training, methods such as Direct Preference Optimization (DPO)~\cite{rafailov2023direct} could leverage structured preferences over HTS classifications (e.g., preferring correct 10-digit codes over near-misses, or valid reasoning traces over hallucinated ones). This would allow the model to learn not just to imitate CROSS rulings but to actively steer away from incorrect classifications. 
\end{itemize}

These directions highlight that while \textsc{Atlas} establishes a strong dense-model baseline, HTS classification remains an open problem with substantial room for methodological innovation.

\section{Results and Evaluation}
\label{sec:results}

We evaluate all models on a held-out test set of 200 CROSS rulings. 
The task requires classifying the correct 10-digit HTS US code for each product description. 
Since tariff classification is inherently hierarchical, we report three complementary evaluation metrics:  

\begin{itemize}
    \item \textbf{Fully correct classification:} all 10 digits match exactly. A fully correct classification means that the end-to-end classification pipeline produces an operationally valid HTS US code, enabling the product to clear customs.  
    \item \textbf{Partially correct classification:} the first 6 digits (harmonized across all WTO members) match. This reflects whether the model generalizes to the globally standardized portion of the code, making it directly relevant for cross-border deployments outside the U.S.  
    \item \textbf{Average digit-level accuracy:} mean number of correctly predicted digits (0–10), capturing fine-grained improvements even when full correctness is not achieved.  
\end{itemize}

\subsection{Fully Correct Classifications}
Table~\ref{tab:full_correct} reports the number and percentage of classifications that exactly match the 10-digit HTS US code. 
General-purpose LLMs such as GPT-5-Thinking achieve moderate success (25\%), whereas open-source baselines without domain adaptation perform poorly ($\leq 3$\%).  
Our fine-tuned model, \textbf{Atlas}, based on LLaMA-3.3-70B, achieves the best results with 40\% fully correct classifications—meaning nearly half of all test products are classified into a customs-ready code.  

\begin{table}[h]
\centering
\begin{tabular}{l|c|c}
\hline
\textbf{Model} & \textbf{Fully Correct (N)} & \textbf{Accuracy (\%)} \\
\hline
GPT-5-Thinking & 50 & 25.0\% \\
Gemini-2.5-Pro-Thinking & 27 & 13.5\% \\
DeepSeek-R1 (05/28) & 5 & 2.5\% \\
GPT-OSS-120B & 3 & 1.5\% \\
LLaMA-3.3-70B & 3 & 2.1\% \\
\textbf{Atlas (Fine-tuned LLaMA-3.3-70B)} & \textbf{80} & \textbf{40.0\%} \\
\hline
\end{tabular}
\caption{Fully correct HTS US code classifications (10-digit match) on the 200-sample test set.}
\label{tab:full_correct}
\end{table}

\subsection{Partially Correct Classification}
Table~\ref{tab:partial_correct} evaluates classifications at the 6-digit level, which is harmonized globally and thus forms the basis for international tariff schedules.  
Here, GPT-5-Thinking reaches 55.5\% accuracy, while \textbf{Atlas} achieves 57.5\%.  
This shows that our domain-specific fine-tuning not only improves U.S.-specific classification but also transfers to the globally harmonized layer, demonstrating potential for adoption in worldwide trade contexts.  

\begin{table}[h]
\centering
\begin{tabular}{l|c|c}
\hline
\textbf{Model} & \textbf{Partially Correct (N)} & \textbf{Accuracy (\%)} \\
\hline
GPT-5-Thinking & 111 & 55.5\% \\
Gemini-2.5-Pro-Thinking & 62 & 31.0\% \\
DeepSeek-R1 (05/28) & 53 & 26.5\% \\
GPT-OSS-120B & 16 & 8.0\% \\
LLaMA-3.3-70B & 29 & 20.7\% \\
\textbf{Atlas (Fine-tuned LLaMA-3.3-70B)} & \textbf{115} & \textbf{57.5\%} \\
\hline
\end{tabular}
\caption{Partially correct HTS code classifications (6-digit harmonized match).}
\label{tab:partial_correct}
\end{table}

\subsection{Average Digit-Level Accuracy}
Finally, we report the average number of digits correctly predicted per code in Table~\ref{tab:avg_digits}.  
While general-purpose models hover around 3–5 digits correct, \textbf{Atlas} achieves 6.3 digits correct on average.  
This demonstrates that supervised fine-tuning on CROSS rulings improves fine-grained reasoning over tariff codes, even when full correctness is not reached.  

\begin{table}[h]
\centering
\begin{tabular}{l|c}
\hline
\textbf{Model} & \textbf{Avg. Digits Correct (out of 10)} \\
\hline
GPT-5-Thinking & 5.61 \\
Gemini-2.5-Pro-Thinking & 2.92 \\
DeepSeek-R1 (05/28) & 3.24 \\
GPT-OSS-120B & 2.58 \\
LLaMA-3.3-70B & 3.31 \\
\textbf{Atlas (Fine-tuned LLaMA-3.3-70B)} & \textbf{6.30} \\
\hline
\end{tabular}
\caption{Average number of correctly predicted digits per HTS code.}
\label{tab:avg_digits}
\end{table}

\subsection{Visual Comparison}
\begin{figure}[!t]
    \centering
    \begin{tikzpicture}
        \begin{axis}[
            ybar=0.4,
            bar width=10pt,
            width=0.9\linewidth,
            height=7cm,
            enlarge x limits=0.18,
            ylabel={Accuracy (\%)},
            symbolic x coords={GPT-5-Thinking,Gemini-2.5-Pro,DeepSeek-R1,GPT-OSS-120B,LLaMA-3.3-70B,Atlas},
            xtick=data,
            x tick label style={rotate=25, anchor=east},
            ymin=0, ymax=65,
            legend style={at={(0.5,1.08)}, anchor=south, legend columns=-1},
            nodes near coords,
            nodes near coords align={vertical},
        ]
        \addplot coordinates {
            (GPT-5-Thinking,25.0)
            (Gemini-2.5-Pro,13.5)
            (DeepSeek-R1,2.5)
            (GPT-OSS-120B,1.5)
            (LLaMA-3.3-70B,2.1)
            (Atlas,40.0)
        };
        \addplot coordinates {
            (GPT-5-Thinking,55.5)
            (Gemini-2.5-Pro,31.0)
            (DeepSeek-R1,26.5)
            (GPT-OSS-120B,8.0)
            (LLaMA-3.3-70B,20.7)
            (Atlas,57.5)
        };
        \legend{10-digit accuracy,6-digit accuracy}
        \end{axis}
    \end{tikzpicture}
    \caption{Visual comparison of model performance on HTS classification. 
    Atlas leads at both evaluation levels, with a marked margin at the 10-digit (U.S.-specific) level.}
    \label{fig:model_comparison}
\end{figure}
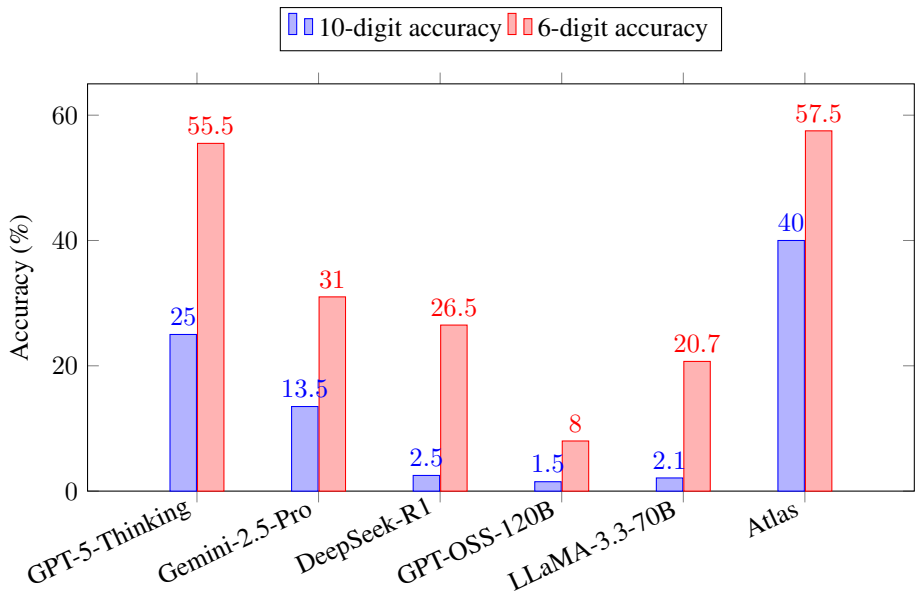

To complement the tables, Figure~\ref{fig:model_comparison} summarizes model performance across both evaluation levels. Atlas’s advantage is especially pronounced at the 10-digit U.S.-specific classification task.  

\subsection{Cost Efficiency of Inference}
Beyond accuracy, cost per inference is a critical factor for practical deployment of tariff classification systems. 
Closed-source API models such as GPT-5-Thinking and Gemini-2.5-Pro-Thinking incur substantial per-query costs, particularly when scaled to thousands of classifications, whereas fine-tuned open-source models can be hosted locally or on cost-effective GPU clusters at a fraction of the price.

Table~\ref{tab:cost_per_thousand} compares the cost of classifying 1,000 product descriptions into 10-digit HTS codes. 
We assume a standard context length ($\sim$1k input tokens, $\sim$200 output tokens) and use publicly available API pricing at the time of writing. 
For open-source models (LLaMA-3.3-70B and Atlas), costs are estimated from on-demand A100 GPU cloud pricing. 

\begin{table}[h]
\centering
\begin{tabular}{l|c}
\hline
\textbf{Model} & \textbf{Cost for 1,000 HTS Inferences (USD)} \\
\hline
GPT-5-Thinking & $\sim\$3.30$ \\
Gemini-2.5-Pro-Thinking & $\sim\$5.50$ \\
DeepSeek-R1 (05/28) & $\sim\$1.00$ \\
GPT-OSS-120B & $\sim\$0.90$ (estimated compute) \\
LLaMA-3.3-70B & $\sim\$0.70$ (self-hosted) \\
\textbf{Atlas (Fine-tuned LLaMA-3.3-70B)} & \textbf{$\sim\$0.70$ (self-hosted)} \\
\hline
\end{tabular}
\caption{Estimated cost of classifying 1,000 products into 10-digit HTS codes. 
Closed-source models use API pricing; open-source models assume on-demand A100 GPU hosting.}
\label{tab:cost_per_thousand}
\end{table}

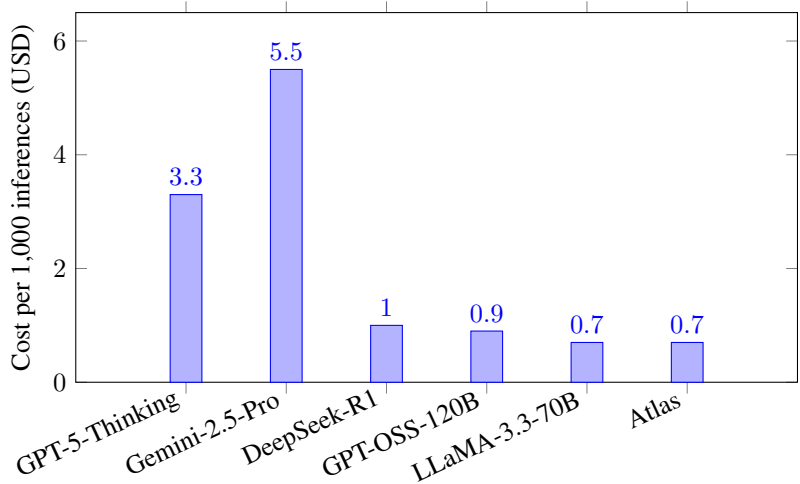
\begin{figure}[!t]
    \centering
    \begin{tikzpicture}
        \begin{axis}[
            ybar,
            bar width=12pt,
            width=0.8\linewidth,
            height=6.5cm,
            enlarge x limits=0.22,
            ylabel={Cost per 1{,}000 inferences (USD)},
            symbolic x coords={GPT-5-Thinking,Gemini-2.5-Pro,DeepSeek-R1,GPT-OSS-120B,LLaMA-3.3-70B,Atlas},
            xtick=data,
            x tick label style={rotate=25, anchor=east},
            ymin=0, ymax=6.5,
            nodes near coords,
            nodes near coords align={vertical},
        ]
        \addplot coordinates {
            (GPT-5-Thinking,3.30)
            (Gemini-2.5-Pro,5.50)
            (DeepSeek-R1,1.00)
            (GPT-OSS-120B,0.90)
            (LLaMA-3.3-70B,0.70)
            (Atlas,0.70)
        };
        \end{axis}
    \end{tikzpicture}
    \caption{Estimated cost per 1{,}000 HTS inferences. Atlas (self-hosted) is substantially cheaper than proprietary APIs.}
    \label{fig:cost_comparison}
\end{figure}

\subsection{Discussion}
Taken together, these results highlight a critical tradeoff:  
\textbf{Atlas} not only surpasses GPT-5-Thinking in accuracy (40\% vs. 25\% fully correct classifications), but also reduces inference cost by nearly \textbf{5$\times$ compared to GPT-5} and almost \textbf{8$\times$ compared to Gemini-2.5-Pro-Thinking}.  
Moreover, the strong performance on partially correct classifications demonstrates that Atlas generalizes beyond U.S.-specific tariffs to the globally harmonized 6-digit regime, reinforcing its utility for international trade applications.

\section{Summary and Future Directions}
\label{summary}

This work introduced the first real world benchmark for Trade policy reasoning based on Harmonized Tariff Schedule (HTS) code classification and presented \textsc{Atlas}, a fine-tuned LLaMA-3.3-70B model adapted to this high-stakes domain. Our study establishes tariff classification as a challenging new frontier for LLM evaluation, with three central takeaways:

\begin{itemize}
    \item \textbf{State-of-the-art performance:} \textsc{Atlas} achieves 40\% fully correct classifications at the 10-digit level and 57.5\% at the 6-digit level, outperforming GPT-5-Thinking (+15 points) and Gemini-2.5-Pro-Thinking (+27.5 points).  
    \item \textbf{Cost and deployment efficiency:} \textsc{Atlas} is nearly \textbf{5$\times$ cheaper than GPT-5} and \textbf{8$\times$ cheaper than Gemini}, while enabling self-hosted deployment for sensitive trade and supply-chain applications.  
    \item \textbf{Open benchmark challenge:} Despite these gains, best 10-digit accuracy remains only 40\%, underscoring the need for advances in reasoning, retrieval, and alignment methods.  
\end{itemize}

Looking forward, we see three promising directions: (1) expanding the dataset to include a broader range of rulings beyond the current subset, (2) distilling \textsc{Atlas} into smaller variants (e.g., 8B or 3B) for efficient deployment in resource-constrained settings, and (3) exploring enhanced reasoning techniques and retrieval-augmented methods to improve classification accuracy.  

We release \textsc{Atlas} \cite{yuvraj2025atlas} and the benchmark splits on Hugging Face \cite{flexify2025huggingface} to support reproducibility. By framing HTS classification as a benchmark task, we aim to catalyze progress on domain-specialized LLMs—directly tied to the resilience of global trade and supply chains.
\bibliography{iclr2026_conference}
\bibliographystyle{iclr2026_conference}

\end{document}